\documentclass[letterpaper]{article}

\usepackage{type1cm}         

\usepackage{makeidx}         % allows index generation
\usepackage{graphicx}        % standard LaTeX graphics tool
\usepackage{multicol}        % used for the two-column index
\usepackage[bottom]{footmisc}% places footnotes at page bottom

\usepackage{newtxtext}       % 
\usepackage{newtxmath}       % selects Times Roman as basic font

\makeatletter
\def\blfootnote{\gdef\@thefnmark{}\@footnotetext}
\makeatother
\usepackage{color}
\usepackage{tabulary}
\usepackage{subcaption}
\usepackage{multirow}
\usepackage[numbers]{natbib}
\makeindex             % used for the subject index
\title{Medication Regimen Extraction From Medical Conversations}

\author{
  Sai P. Selvaraj, Sandeep Konam\\
  Abridge AI Inc.\\
  {\it \{prabhakarsai, san\}@abridge.com}
}
\date{} % remove this

\begin{document}
\maketitle

\begin{abstract}
Extracting relevant information from medical conversations and providing it to doctors and patients might help in addressing doctor burnout and patient forgetfulness.
In this paper, we focus on extracting the Medication Regimen (dosage and frequency for medications) discussed in a medical conversation. 
We frame the problem as a Question Answering (QA) task and perform comparative analysis over: a QA approach, a new combined QA and Information Extraction approach, and other baselines. We use a small corpus of 6,692 annotated doctor-patient conversations for the task. Clinical conversation corpora are costly to create, difficult to handle (because of data privacy concerns), and thus scarce. We address this data scarcity challenge through data augmentation methods, using publicly available embeddings and pretrain part of the network on a related task (summarization) to improve the model's performance. Compared to the baseline, our best-performing models improve the dosage and frequency extractions' ROUGE-1 F1 scores from 54.28 and 37.13 to 89.57 and 45.94, respectively. Using our best-performing model, we present the first fully automated system that can extract Medication Regimen tags from spontaneous doctor-patient conversations with about $\approx$71\% accuracy.
\end{abstract}

% Our best model can correctly detect the dosage and frequency for 71.75\% and 73.58\% of medications from the transcripts of commercially available (Google and IBM) Automatic Speech Recognition services.

\section{Introduction}
\blfootnote{\hspace{-.72cm} Proceedings of International Workshop on Health Intelligence (W3PHIAI) of the
34\textsuperscript{th} AAAI Conference on Artificial Intelligence, 2020.} % -remove this

Physician burnout is a growing concern, estimated to be experienced by at least 35\% of physicians in the developing world and 50\% in the United States \cite{kumar2016burnout}. It is found that for every hour physicians provide direct clinical facetime to patients, nearly two additional hours are spent on EHR (Electronic Health Records) and administrative or desk work. As per the study conducted by Massachusetts General Physicians Organization (MPGO) \cite{del2019trends} and as reported by \citeauthor{hcirajiv}, the average time spent on administrative tasks increased from 23.7\% in 2014 to 27.9\% in 2017. Both surveys found that time spent on administrative tasks was positively associated with higher likelihood of burnout. Top reasons under administrative burden include working on the ambulatory EHR, handling medication reconciliation (sometimes done by aides), medication renewals, and medical billing and coding. The majority of these reasons revolve around the documentation of information exchanged between doctors and patients during clinical encounters. Automatically extracting such medical information \cite{finley2018automated} could both alleviate the documentation burden on the physician, and also allow them to dedicate more time directly with patients.

Among all the medical information extraction tasks, Medication Regimen (medication name, dosage, and frequency) extraction is particularly interesting due to its ability to help doctors with medication orders cum renewals, medication reconciliation, verifying of reconciliations for errors, and other medication-centered EHR documentation tasks. In addition, the same information when provided to patients can help them better recall doctor's instructions which might aid in compliance with the care plan. This is particularly important given that patients forget or wrongly recollect 40-80\% \cite{kessels2003patients} of what is discussed in the clinic, and accessing EHR data has its own challenges \cite{gao2018}.

Spontaneous medical conversations happening between a doctor and a patient have several distinguishing characteristics from a normal monologue or prepared speech: it involves multiple speakers with overlapping dialogues, covers a variety of speech patterns, and the vocabulary can range from colloquial to complex domain-specific language. With recent advancements in Conversational Speech Recognition \cite{kim2019cross} rendering the systems less prone to errors, the subsequent challenge of understanding and extracting relevant information from the conversations is receiving increased research focus \cite{finley2018automated,kannan2018semi,jeblee2019extracting,liu2019fast}.

In this paper, we focus on \emph{local} information extraction in transcribed medical conversations. Specifically, we extract dosage (e.g. 5mg) and frequency (e.g. once a day) for the medications (e.g. aspirin) from these transcripts, collectively referred to as \emph{Medication Regimen (MR) extraction}. The information extraction is \emph{local} as we extract the information from a segment of the transcript and not the entire transcript since doing the latter is difficult owing to the long meandering nature of the conversations often with multiple medication regimens and care plans being discussed. 

The challenges associated with the Medication Regimen (MR) extraction task include understanding the spontaneous dialog with medical vocabulary and understanding the relationship between different entities as the discussion can contain multiple medications and dosages (e.g. doctor revising a dosage or reviewing all the medications).

We frame this problem as a Question Answering (QA) task by generating questions using templates. We base the QA model on pointer-generator networks \cite{see2017get} augmented with Co-Attentions \cite{xiong2016dynamic}. In addition, we develop models combining QA and Information Extraction frameworks using multi-decoder (one each for dosage and frequency) architecture. %We also compare our model with baseline models and trained an abstractive QA model trained on the Decathlon challenge  \cite{McCann2018decaNLP} -- MQAN. 

Lack of availability of a large volume of data is a typical challenge in healthcare. A conversation corpus by itself is a rare commodity in the healthcare data space because of the cost and difficulty in handing (because of data privacy concerns). Moreover, transcribing and labeling the conversations is a costly process as it requires domain-specific medical annotation expertise. To address data shortage and improve the model performance, we investigate different high-performance contextual embeddings (ELMo \cite{peters2018deep}, BERT \cite{devlin2019bert} and ClinicalBERT \cite{alsentzer-etal-2019-publicly}), and pretrain models on a medical summarization task. We further investigate the effects of training data size on our models.

On the MR extraction task, ELMo with encoder multi-decoder architecture and BERT with encoder-decoder with encoders pretrained on the summarization task perform the best. The best-performing models improve our baseline's dosage and frequency extractions ROUGE-1 F1 scores from 54.28 and 37.13 to 89.57 and 45.94, respectively.

Using our models, we present the first fully automated system to extract MR tags from spontaneous doctor-patient conversations. We evaluate the system (using our best performing models) on the transcripts generated from Automatic Speech Recognition (ASR) APIs offered by Google and IBM. In Google ASR's transcripts, our best model obtained ROUGE-1 F1 of 71.75 for Dosage extraction (which in this specific case is equal to the percentage of times dosage is correct, refer to Section: \ref{sec:metrics} for more details) and 40.13 for Frequency extraction tasks. On qualitative evaluation, we find that for 73.58\% of the medications the model can find the correct frequency.
%And, the best model can find the correct dosage for 71.75\% ROUGE-1 F1 in this case see Section Metrics), and the frequency for 73.58\% of the medications (that were recognized correctly) from the Google ASR transcripts.
These results demonstrate that the research on NLP can be used effectively in a real clinical setting to benefit both doctors and patients.

\begin{table}[!t]
    \centering
    \tiny
    \begin{tabulary}{\linewidth}{|p{4cm}|p{3.2cm}|p{.82cm}|C|C|}
        \hline
         (Timestamp) Transcript & Summary & \texttt{Medication Name} & \texttt{Dosage} & \texttt{Frequency} \\ 
        \hline
        (1028.3 s) So, let's, I'm going to have them increase the mg, uh, Coumadin level, so that, uh, like I said, the pulmonary embolism doesn't get worse here. & Increase Coumadin to 3.5 mg to prevent pulmonary embolism from getting bigger & Coumadin & 3.5 mg & Twice a day \\
        \hfill (1044.9 s) Yeah.  & & & & \\
        (1045.2 s) Yeah. & & & & \\
        \hspace{1.2 cm} (1045.4 s) Increase it to three point five& & & &\\
        \multicolumn{1}{|r|}{in the morning and and before bed.} & & & &\\
    \hline
    \end{tabulary}
\caption{Example of our annotations grounded to the transcript segment.}
\label{tab:data}
\end{table}
        % \hline
        % (29.1 s) I'll, I'll send it to the clinic, Coumadin, yeah, Aspirin, but, again, baby aspirin I want you on 120. & Baby aspirin 120 mg &  Baby aspirin & 120 mg & Once a day  \\
        % \hfill (35.4 s) Yeah, but I am on 60 mg daily.  & & & & \\
        % (37.3 s) I want you to be on 120 now. &  & &  & \\

\section{Data}

Our dataset consists of a total of 6,693 real doctor-patient conversations recorded in a clinical setting using distant microphones of varying quality. The recordings have an average duration of 9min 28s and have a verbatim transcript of 1,500 words on average (written by the experts). Both the audio and the transcript are de-identified (by removing the identifying information) with digital zeros and \texttt{[de-identified]} tags, respectively. The sentences in the transcript are grounded to the audio with the timestamps of their first and last words.

The transcript of the conversations are annotated with summaries and Medication Regimen tags (MR tags), both grounded using the timestamps of the sentences from the transcript deemed relevant by the expert annotators, refer to Table \ref{tab:data}.
The transcript for a typical conversation can be quite long, and not easy for many of the high performing deep learning models to act on. Moreover, the medical information about a concept/condition/entity can change during the conversation after a significant time gap. For example, the dosage of a medication can be different when discussing current medication the patient is on, versus when they are prescribed a different dosage. For this reason, we have annotations, that are grounded to a short segment of the transcript.

The summaries (\#words - $\mu$=9.7; $\sigma$=10.1) are medically relevant and local.
The MR tags are also local and are of the form \{\texttt{Medication Name}, \texttt{Dosage}, \texttt{Frequency}\}. If dosage ($\mu$=2.0; $\sigma$=0) or frequency ($\mu$=2.1; $\sigma$=1.07) information for a medication is not present in a grounded sentence, the corresponding field will be `none'.

In the MR tags, \texttt{Medication Name} and \texttt{Dosage} (usually a quantity followed by its units) can be \textit{extracted} with relative ease from the transcript except for the units of the dosage, which is sometimes inferred. In contrast, due to high degree of linguistic variation with which \texttt{Frequency} is often expressed, extracting it requires an additional \textit{inference} step. For example, `take one in the morning and at noon' from the transcript is tagged as `twice a day' in the frequency tag, likewise `take it before sleeping' is tagged as `at night time'.

Out of 6,693 files, we set aside a random sample of 423 files (denoted as $\mathcal{D}_{test}$) for final evaluation.
The remaining 6,270 files are considered for training with 80\% train (5016), 10\% validation (627), and 10\% test (627) split. Overall, the 6,270 files contain 156,186 summaries and 32,000 MR tags, out of which 8,654 MR tags contain values for at least one of the \texttt{Dosage} or \texttt{Frequency}, which we used for training to avoid overfitting (the remaining MR tags have both \texttt{Dosage} and \texttt{Frequency} as `none'). Note that we have two test datasets: `10\% test' - used to evaluate all the models, and $\mathcal{D}_{test}$ - used to measure the performance of best performing models on ASR transcripts.

\section{Approach}

We frame the Medication Regimen extraction problem as a Question Answering (QA) task, which forms the basis for our first approach. It can also be considered as a specific inference or relation extraction task, since we extract specific information about an entity (\texttt{Medication Name}). For this reason, our second approach is at the intersection of Question Answering (QA) and Information Extraction (IE) domains. Both approaches involve using a contiguous segment of the transcript and the \texttt{Medication Name} as input to find or infer the medication's \texttt{Dosage} and \texttt{Frequency}. When testing the approaches mimicking real-world conditions, we extract \texttt{Medication Name} from the transcript separately using medication ontology (refer to Section: \ref{sec:eval_asr}).

In the first approach, we frame the MR task as a QA task and generate questions using the template: ``What is the  $ <$dosage/frequency$>$ for $<$\texttt{Medication Name}$>$?". Here, we use an abstractive QA model based on pointer-generator networks \cite{see2017get} augmented with coattention encoder \cite{xiong2016dynamic} (QA-PGNet).

In the second approach, we frame the problem as a conditioned IE task, where the information extracted depends on an entity (\texttt{Medication Name}). Here, we use a multi-decoder pointer-generator network augmented with coattention encoder (Multi-decoder QA-PGNet). Instead of using templates to generate questions and single decoder to extract different types of information as in the QA approach (which might lead to performance degradation), here we consider separate decoders for extracting specific types of information conditioned on an entity $E$ (\texttt{Medication Name}).

\subsection{Pointer-generator Network (PGNet)}

The network is a sequence-to-sequence attention model that can both copy a word from the input $I$ containing $P$ word tokens or generate a word from its vocabulary $vocab$, to produce the output sequence.

First, embeddings of the input tokens are fed one-by-one to the encoder, a single bi-LSTM layer which encodes the tokens into hidden states - $H = encoder(I)$, where $H = [h_1...h_P]$.
For each decoder time step $t$, in a loop, we compute, 1) attention $a_t$ (using the last decoder state $s_{t-1}$), over the input, and 2) the decoder state $s_t$ using $a_t$. Then, at each time step, using both $a_t$ and $s_t$ we can find the probability $P_t(w)$, of producing a word $w$ (from both $vocab$ and $I$). For convenience, we denote the attention and the decoder as $decoder_{pg}(H) = P(w)$, where $P(w) = [P_1(w)...P_T(w)]$. The output can then be decoded from $P(w)$, which is decoded until it produces an `end of output token' or the number of steps reach the maximum allowed limit.

\subsection{QA PGNet}
We first separately encode both the question - $H_Q$=$encoder(Q)$ and the input - $H_I=encoder(I)$ using encoders (with shared weights). Then, to condition $I$ on $Q$ (and vice versa), we use the coattention encoder \cite{xiong2016dynamic} which attends to both the $I$ and $Q$ simultaneously to generate the coattention context - $C_D$=$coatt(H_I, H_Q)$. Finally, using the pointer-generator decoder we find the probability distribution of the output sequence - $P(w) = decoder_{pg}([H_I; C_D])$, which is then decoded to generate the answer sequence.

\subsection{Multi-decoder (MD) QA PGNet}

For extracting $K$ types of information about an entity $E$, we first encode the inputs into $H_I$ and $H_E=encoder(E)$. Then in an IE fashion, we use multi-decoder (MD) setup to obtain $K$ probability distributions $P^k(w)$, which can then be decoded to get the corresponding output sequences.
$$ C_D^k = coatt^k(H_I, H_E)\  \forall k=1...K $$
$$ P^k(w) = decoder_{pg}^k([H_I; C_D^k])\  \forall k=1...K $$
% Predictions for each of the $K$ decoders are then decoded using $P^k(w)$.

All the networks discussed above are trained using a negative log-likelihood loss.
% for the target word at each time step and summed over all the decoder time steps.

\section{Experiments}

We initialized MR extraction models' vocabulary from the training dataset after removing words with a frequency lower than 30 in the dataset, resulting in 456 word tokens. Our vocabulary is small because of the size of the dataset, hence we rely on the model's ability to copy words to produce the output effectively. In all our model variations, the embedding and the network's hidden dimension are set to be equal. The networks were trained with a learning rate of 0.0015, dropout of 0.5 on the embedding layer, normal gradient clipping set at 2, batch size of 8, and optimized with Adagrad and the training was stopped using the $10\%$ validation dataset.
%  \cite{duchi2011adaptive}

\subsection{Data Processing}

We did the following basic preprocessing to our data: 1) added `none' to the beginning of the input utterance so that the network could point to it when there was no relevant information in the input, 2) filtered outliers with a large number of grounded transcript sentences ($>$150 words),  and 3) converted all text to lower case.

To improve performance, we 1) standardized all numbers (both digits and words) to words concatenated with a hyphen\footnote{This prevents overfitting and repetition when converting all the numbers to words.}  (e.g. 110 -$>$ one-hundred-ten), in both input and output, 2) removed units from \texttt{Dosage} as sometimes the units are not explicitly mentioned in the transcript segment but were written by the annotators using domain knowledge, 3) prepended all medication mentions with `rx-' tag, as this helps the model's performance when multiple (different) medications are discussed in a segment, and 4) created new data points by randomly shuffling medications and dosages in both input and output (when we have more than one in a transcript segment) to increase the number of training data points. Randomly shuffling the entities increases the number of training MR tags from 8,654 to 11,521. Based on the data statistics after data processing, we fixed the maximum encoder steps to 100, dosage decoder steps to 1, and frequency decoder steps to 3 (for both the QA and Multi-decoder QA models).

\subsection{Metrics}
\label{sec:metrics}

For the MR extraction task, we measure the ROUGE-1 scores for both the Dosage and Frequency extraction tasks. It should be noted that since \texttt{Dosage} is a single word token (after processing), both the reference and hypothesis are a single token, making its ROUGE-1 F1, Precision and Recall scores equal, which are in turn equal to the percentage of times we find the correct dosage for the medications.

In our annotations, Frequency has conflicting tags (e.g. \{`Once a day', `twice a day'\} and `daily'), hence metrics like Exact Match will be erroneous. To address this issue, we use the ROUGE scores to compare different models on the 10\% test dataset and we use qualitative evaluation to measure the top-performing models on $\mathcal{D}_{test}$.

\begin{table}[!htb]
    \centering
    \begin{tabular}{|c||c||c|c|c|}
        \hline
        \multirow{2}{*}{Models} & Dosage & \multicolumn{3}{c|}{Frequency}\\
        \cline{2-5}
        & F1 & F1 & Recall & Precision \\
        \hline
        Nearest Number (B)  & 67.13 & - & - & - \\
        Random Top-3 (B) & - & 29.32 & 29.20 & 29.49 \\
        \hline
        Lookup table + QA PGNet (B) & 54.28 & 37.13 & 59.83 & 29.82 \\
        ELMo + QA PGNet & 88.67 & 35.04 & 59.94 & 26.91 \\
        BERT + QA PGNet & 86.34 & 43.52 & 70.46 & 34.81 \\
        ClinicalBERT + QA PGNet & 84.20 & 43.92 & 73.67 & 34.27  \\
        \hline
        Lookup table + MD QA PGNet (B) & 51.97 & 36.56 & 71.37 & 25.00 \\
        ELMo + MD QA PGNet & 88.64 & 37.41 & 58.66 & 30.56 \\
        BERT + MD QA PGNet & 85.59 & 42.57 & 71.05 & 33.33 \\
        ClinicalBERT + MD QA PGNet & 84.82 & 44.04 & 71.50 & 33.05 \\
        \hline
%        Lookup table + QA PGNet(PT)& & & & \\
        ELMo + QA PGNet(PT)& 89.17 & 42.78 & 70.61 & 33.54 \\
        BERT + QA PGNet(PT)& 88.62 & \bf{45.94} & 74.54 & 36.70 \\
%        ClinicalBERT + QA PGNet(PT)& & & & \\
        \hline
%       Lookup table + Multi-decoder QA PGNet(PT)& & & & \\
        ELMo + MD QA PGNet(PT) & \bf{89.57} & 44.82 & 75.71 & 34.62 \\
        BERT + MD QA PGNet(PT) & 86.98 & 44.47 & 74.43 & 34.75 \\
%        ClinicalBERT + Multi-decoder QA PGNet(PT) & & & & \\
        \hline
    \end{tabular}
  \caption{ROUGE-1 scores of baselines and models for the MR extraction task on the 10\% test dataset. PT: using pretrained encoder; B: baseline; MD: Multi-decoder }
  \label{tab:main-r}
\end{table}

\subsection{Model variations}

We consider QA PGNet and Multi-decoder QA PGNet with lookup table embedding as baseline models and improve over the baselines with other variations described below. Apart from learning-based baselines, we also create two naive baselines. For Dosage extraction, the baseline we consider is `Nearest Number', where we take the number nearest to the \texttt{Medication Name} as the prediction, and `none' if none exist or if the \texttt{Medication Name} is not detected in the input\footnote{This can happen when a different form of a medication (e.g. abbreviation, generic or brand name) is used in the conversation compared to the annotation}. For Frequency extraction, the baseline we consider is `Random Top-3' where we predict a random \texttt{Frequency} tag, from top-3 most frequent ones from our dataset - \{`none', `daily', `twice a day'\}.

{\bf \noindent Embedding:}
We developed different variations of our models with a simple lookup table embeddings learned from scratch and using high-performance contextual embeddings, which are ELMo \cite{peters2018deep}, BERT \cite{vaswani2017attention} and ClinicalBERT \cite{alsentzer-etal-2019-publicly} (trained and provided by the authors). Refer to Table \ref{tab:main-r} for the performance comparisons.

We derive embeddings from ELMo by learning a linear combination of its last three layer's hidden states (task-specific fine-tuning \cite{peters2018deep}). Similarly, for BERT-based embeddings, we take a linear combination of the hidden states from its last four layers, as this combination performs best without increasing the size of the embeddings \cite{vaswani2017attention}. Since BERT and ClinicalBERT use word-piece vocabulary and compute sub-word embeddings, we compute word-level embedding by averaging the corresponding sub-word tokens. ELMo and BERT embeddings both have 1024 dimensions, ClinicalBERT have 768 as it is based on BERT base model, and the lookup table have 128 -- higher dimension models leads to overfitting. %The corresponding network's hidden dimension that we train (with these embeddings) is the same as the embeddings.

{\bf \noindent Pertaining Encoder:} We trained the PGNet as a summarization task using the medical summaries and used the trained model to initialize the encoders (and the embeddings) of the corresponding QA models. We use a vocab size of 4073 words, derived from the training dataset with a frequency threshold of 30 for the task. We trained the models using Adagrad optimizer with a learning rate of 0.015, normal gradient clipping set at 2 and trained for around 150000 iterations (stopped using validation dataset). On the summarization task PGNet obtained ROUGE-1 F1 scores of 41.42 with ELMo and 39.15 with BERT embeddings. We compare the effects of pretraining the model in Table: \ref{tab:main-r}, models with `pretrained encoder' had their encoders and embeddings pretrained with the summarization task.
% \footnote{It should be noted that when we are pretraining models with contextual embeddings we do not have to maintain the vocabulary between the two tasks.}.
%On the other hand, it is necessary to maintain the vocabulary when pretraining the models with lookup table embeddings.
% Hence our model with pretrained lookup table PGNet has a vocabulary of 4073 words.

\section{Results and Discussion}

\subsection{Difference in networks and approaches}

{\bf \noindent Embeddings: } On Dosage extraction, in general, ELMo obtains better performance  than BERT, refer to Table \ref{tab:main-r}. This could be because we concatenated the numbers with a hyphen, and because ELMo uses character-level tokens it can learn the tagging better than BERT. Similar observations are found in recent literature. On the other hand, on Frequency extraction, without pretraining, ELMo's performance lags by a big margin of $\approx$8.5 ROUGE-1 F1 compared to BERT-based embeddings.
% \citeauthor{wallace2019nlp}

Although ClinicalBERT performed the best in the Frequency extraction task (by a small margin) in cases without encoder pretraining, in general it does not perform as well as BERT. This could also be a reflection of the fact that the language and style of writing used in clinical notes is very different from the way doctors converse with patients and the embedding dimension difference. Lookup table embedding performed decently in the frequency extraction task, but lags behind in the Dosage extraction task.

%From the metrics and qualitative inspection, we find that the Frequency extraction is an easier task than the Dosage extraction. This is because medical conversations tend to have frequency information occur in isolation and near the medications, while a medication's dosage can occur 1) near other medication's dosages, 2) with previous dosages (when a dosage for a medication is revised), and 3) after a large number of words from the medication.

{\bf \noindent Other Variations:}
Considering various models' performance (without pretraining) and the resource constraint, we choose ELMo and BERT embeddings to analyze the effects of pretraining the encoder. When the network's encoder (and embedding) is pretrained with the summarization task, we 1) see a small decrease in the average number of iterations required for training, 2) improvement in individual performances of all models for both the sub-tasks, and 3) get best performance metrics across all variations, refer to Table \ref{tab:main-r}. Both in terms of performance and the training speed, there is no clear winner between shared and multi-decoder approaches. Medication tagging and data augmentation increase the best-performing model's ROUGE-1 F1 score by $\approx$1.5 for the Dosage extraction task.
%However, no significant difference was observed in the Frequency extraction task.

We also measure the performance of Multitask Question Answering Network (MQAN) \cite{mccann2018natural} a QA model trained by the authors on the Decathlon multitask challenge. Since MQAN was not trained to produce our output sequence, it would not be fair to compute ROUGE scores, hence we haven't included them in the tables. Instead, we randomly sample the MQAN's predictions from the 10\% test dataset and qualitatively evaluate it. From the evaluations, we find that MQAN can not distinguish between frequency and dosage, and mixed the answers. MQAN correctly predicted the dosage for 29.73\% and frequency for 24.24\% percent of the medications compared to 84.12\% and 76.34\% for the encoder pretrained BERT QA PGNet model trained on our dataset. This could be because of differences in the training dataset, domain, and the tasks in the Decathlon challenge compared to ours.

Almost all our models perform better than the naive baselines and the ones using lookup table embeddings, and our best performing models outperform them significantly. Among all the variations, the best performing models are ELMo with Multi-decoder (Dosage extraction) and BERT with shared-decoder QA PGNet architecture (Frequency extraction) with pretrained encoder. We chose these two models for our subsequent analysis.

\subsection{Breakdown of Performance}
\label{sec:per-break}

We categorize the 10\% test dataset into different categories based on the complexity and type of the data and analyze the breakdown of the system's performance in Table \ref{tab:per-breakdown}. We breakdown the Frequency extraction into two categories,: 1) None: ground truth Frequency tag is `none', and 2) NN (Not None): ground truth Frequency tag is not `none'. Similarly, the Dosage extraction into 4 categories: 1) None: ground truth dosage tag is `none', 2) MM (Multiple Medicine): input segment has more than one Medication mentioned, 3) MN (Multiple Numbers): input segment has more than one number present, and 4) NBM (Number between correct Dosage and Medicine) : between the \texttt{Medication Name} and the correct Dosage in the input segment there are other numbers present. Note that the categories of the Dosage extraction task are not exhaustive, and one tag can belong to multiple categories.

\begin{table}[!tb]
    \centering
    \begin{tabular}{|c||c|c|c|c||c|c|}
        \hline
        \multirow{2}{*}{Models} & \multicolumn{4}{c||}{Dosage} & \multicolumn{2}{c|}{Frequency}\\
        \cline{2-7}
        & None & MM & MN & NBM & None & NN \\
        \hline
        ELMo + MD QA PGNet                       & \bf{93.46} & 81.73 & 69.87 & \bf{66.68} & 42.98 & 34.58 \\
        BERT + QA PGNet                                     & 93.12 & 78.38 & 65.51 & 53.32  & 45.10 & 42.13 \\
        \hline
         ELMo + MD QA PGNet(PT)& 92.29 & \bf{84.17} & \bf{74.01}  & 60.02 & \bf{45.77} & 45.02\\
        BERT + QA PGNet(PT)               & \bf{93.46} & 79.95 & 68.77  & 60.02 & 43.31 & \bf{46.11}\\
        \hline
    \end{tabular}
   \caption{Performance (ROUGE-1 F1) breakdown of the best performing models measured on the 10\% test dataset for the MR extraction task, refer Results section for more details. PT: using pretrained encoder; MD: Multi-decoder}
   \label{tab:per-breakdown}
\end{table}

From the performance breakdown of Dosage extraction task, we see that 1) the models are able to better identify when a medication's dosage is absent (`none') than other categories, 2) there is a performance dip in hard cases (MM, MN, and NBM), 3) the models are able to figure out the correct dosage (decently) for a medication even when there are multiple numbers/dosage present, and 4) the model struggles the most in the NBM category. The models' low performance in NBM could be because we have a comparatively lower number of examples to train in this category. The Frequency extraction task performs equally well when the tag is `none` or not. In most categories, we see an increase in performance when using pretrained encoders.

\begin{figure}
     \centering
     \begin{subfigure}[b]{0.49\textwidth}
         \centering
         \includegraphics[width=\textwidth]{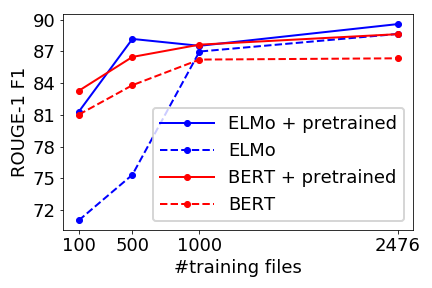}
         \caption{Dosage Extraction}
     \end{subfigure}
     \hfill
     \begin{subfigure}[b]{0.49\textwidth}
         \centering
         \includegraphics[width=\textwidth]{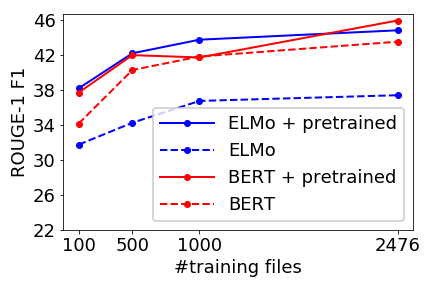}
         \caption{Frequency Extraction}
     \end{subfigure}
 \caption{Difference in the performance of the models on different training data sizes (100, 500, 1,000, and 2,476) on the 10\% test dataset for the MR extraction task.}
    \label{fig:datasize}
\end{figure}

\subsection{Training Dataset Size}

We vary the number of MR tags used to train the model and analyze the model's performance when training the networks using publicly available contextual embeddings, compared to using pretrained embeddings and encoder (pretrained on the summarization task). Out of the 5,016 files in the 80\% train dataset, only 2,476 have at least one MR tag. Therefore, out of the 2476 files, we randomly choose 100, 500, and 1000 files and trained the best performing model variations to observe the performance differences (refer to Figure \ref{fig:datasize}). For all these experiments we used the same vocabulary size (456), the same hyper/training parameters, and the same 10\% test split of 627 files.

As expected, we see that the encoder pretrained models have higher performance across all the training data sizes, when compared to their non-pretrained counterparts (refer to Figure \ref{fig:datasize}). The difference, as expected, shrinks as the training data size increases.

\subsection{Evaluating on ASR transcripts}
\label{sec:eval_asr}

To test the performance of our models in real-world conditions, we use commercially available ASR services from Google and IBM to transcribe the $\mathcal{D}_{test}$ files and measure the performance of our models without assuming any annotations (except when calculating the metrics). It should be noted that this is not the case in our previous evaluations using `10\% test' dataset where we use the segmentation information. For ground truth annotations on ASR transcripts, we aligned the MR tags from human written transcripts to the ASR transcript using their grounded timing information. Additionally, since ASR is prone to errors, if a medication from an MR tag is not recognized correctly in the ASR transcript, during the alignment we remove the corresponding MR tag.

In our evaluations, we use Google Cloud Speech-to-Text\footnote{https://cloud.google.com/speech-to-text/} (G-STT) and IBM Watson Speech to Text\footnote{https://www.ibm.com/cloud/watson-speech-to-text} (IBM-STT) as these were among the top-performing ASR APIs on medical speech \cite{kodish2018systematic} and were readily available to us. We used G-STT, with the `video model' with punctuation settings. Unlike our human written transcripts, the transcript provided by G-STT is not verbatim and does not have disfluencies. IBM-STT, on the other hand, does not give punctuation so we used the speaker changes to add end-of-sentence punctuation.

In our $\mathcal{D}_{test}$ dataset, on initial study we see a Word Error Rate of $\approx$50\% for the ASR APIs. This number is not accurate because of, 1) de-identification, 2) disfluency (verbatim) differences between the human written and ASR transcript, and 3) minor alignment differences between the audio and the ground truth transcript.

\begin{table}[tb!]
    \centering
    \begin{tabular}{|c|c||c||c|c|c|}
        \hline
        \multirow{2}{*}{Transcripts} &  \multirow{2}{*}{Models} & Dosage & \multicolumn{3}{c|}{Frequency}\\
        \cline{3-6}
        & & F1 & F1 & R & P \\
        \hline
        HW  & BERT + QA PGNet(PT) & 85.69 & 49.25 & 72.25 & 41.20 \\
        with HS & ELMo + MD QA PGNet(PT) & 85.79  & 43.99 & 69.27 & 35.63 \\
        \hline
        \hline
        HW & BERT + QA PGNet(PT)& 75.43 & 41.62 & 62.29 & 35.01 \\
        with AS & ELMo + MD QA PGNet(PT)& 75.71 & 37.83 & 61.41 & 29.84 \\
        \hline
        \hline
         G-STT & BERT + QA PGNet(PT)& 70.51 & 39.90 & 61.32 & 33.07 \\
        with AS & ELMo + MD QA PGNet(PT)& 71.75 & 40.13 & 67.22 & 31.14 \\
        \hline
        IBM-STT & BERT + QA PGNet(PT) & 73.93 & 30.90 & 52.53 & 24.10 \\
        with AS & ELMo + MD QA PGNet(PT) & 78.93 & 36.58 & 67.03 & 26.52 \\
        \hline
    \end{tabular}
   \caption{ROUGE-1 scores of the best performing models on the ASR (Google: G-STT and IBM: IBM-STT) and human written transcripts (HW) on the $\mathcal{D}_{test}$ dataset for the MR extraction task. PT: using pretrained encoder; HS: Human segmentation; AS: Auto segmentation; MD: Multi-decoder; P: Precision; R: Recall}
   \label{tab:asr_and_hum}
\end{table}

During this evaluation, we followed the same preprocessing methods we used during training. Then, we auto segment the transcript into small contiguous segments similar to the grounded sentences in the annotations for tags extraction. To segment the transcript, we follow a simple procedure. First, we detect all the medications in a transcript using RxNorm \cite{liu2005rxnorm} via string matching\footnote{Since we had high quality human written transcripts and our ASR transcripts did not contain spelling mistakes (as long as the word was correctly recognized), string matching worked well during testing.}. For all the detected medications, we selecte $2 \leq x \leq 5$ nearby sentences as the input to our model. We increased $x$ iteratively until we encounter a quantity entity -- detected using spaCy's entity recognizer\footnote{https://spacy.io/api/entityrecognizer}, and we set $x$ as 2 if we did not detect any entities in the range.

We show the model's performance on ASR transcripts and human written transcripts with automatic segmentation, and human written transcripts with human (defined) segmentation, in Table \ref{tab:asr_and_hum}. Since the number of recognized medications in IBM-STT is only 95 compared to 725 (human written), we mainly consider the models' performance on G-STT's transcripts (343).

On the Medications that were recognized correctly, the models can perform decently on ASR transcripts in comparison to human transcripts (within 5 points ROUGE-1 F1 for both tasks, refer to Table \ref{tab:asr_and_hum}). This shows that the models are robust to ASR variations discussed above. The lower performance compared to human transcripts is mainly due to incorrect recognition of \texttt{Dosage} and other medications in the same segments (changing the meaning of the text). By comparing the performance of the model on the human written transcripts with human (defined) segmentation and the same with auto segmentation, we see a 10 point drop in Dosage and 6 point drop in Frequency extraction tasks. This points out the need for more sophisticated segmentation algorithms.

With G-STT, our best model obtained ROUGE-1 F1 of 71.75 (which equals to percentage of times dosage is correct in this case) for Dosage extraction and 40.13 for Frequency extraction tasks. To measure the percentage of times the correct frequency was extracted by the model, we qualitatively compared the extracted and predicted frequency. We find that the model can find the correct frequency from the transcripts for 73.58\% of the medications.

\section{Conclusion}

In this paper, we explore the Medication Regimen (MR) extraction task of extracting dosage and frequency for the medications mentioned in a doctor-patient conversation transcript. We explore different variations of abstractive QA models and a new architecture at the intersection of QA and IE frameworks and provide a comparative performance analysis of the methods along with other techniques like pretraining to improve the performance. Finally, we demonstrate the performance of our best-performing models by automatically extracting MR tags from spontaneous doctor-patient conversations (using commercially available ASR). Our best model can correctly extract the dosage for 71.75\% (interpretation of ROUGE-1 score) and frequency for 73.58\% (on qualitative evaluation) of the medications discussed in the transcripts generated using Google Speech-To-Text. In summary, we demonstrate that our research can be translated into real clinical settings to realize its benefits for both doctors and patients.

Using ASR transcripts in training to improve the performance and extracting other important medical information can be interesting lines of future work.

\section{Acknowledgements}

We thank: University of Pittsburgh Medical Center (UPMC) and Abridge AI Inc. for providing access to the de-identified data corpus; Dr. Shivdev Rao, CEO, Abridge AI Inc. and a practicing cardiologist in UPMC's Heart and Vascular Institute, and Prlof. Florian Metze, Associate Research Professor, Carnegie Mellon University for helpful discussions; Ben Schloss, Steven Coleman, and Deborah Osakue for data business development and annotation management.

\bibliographystyle{spbasic}
\bibliography{refyui}

\begin{thebibliography}{19}
\providecommand{\natexlab}[1]{#1}
\providecommand{\url}[1]{{#1}}
\providecommand{\urlprefix}{URL }
\expandafter\ifx\csname urlstyle\endcsname\relax
  \providecommand{\doi}[1]{DOI~\discretionary{}{}{}#1}\else
  \providecommand{\doi}{DOI~\discretionary{}{}{}\begingroup
  \urlstyle{rm}\Url}\fi
\providecommand{\eprint}[2][]{\url{#2}}

\bibitem[{Alsentzer et~al.(2019)Alsentzer, Murphy, Boag, Weng, Jindi, Naumann,
  and McDermott}]{alsentzer-etal-2019-publicly}
Alsentzer E, Murphy J, Boag W, Weng WH, Jindi D, Naumann T, McDermott M (2019)
  Publicly available clinical {BERT} embeddings. In: Proceedings of the 2nd
  Clinical Natural Language Processing Workshop, Association for Computational
  Linguistics, Minneapolis, Minnesota, USA, pp 72--78

\bibitem[{del Carmen et~al.(2019)del Carmen, Herman, Rao, Hidrue, Ting,
  Lehrhoff, Lenz, Heffernan, and Ferris}]{del2019trends}
del Carmen MG, Herman J, Rao S, Hidrue MK, Ting D, Lehrhoff SR, Lenz S,
  Heffernan J, Ferris TG (2019) Trends and factors associated with physician
  burnout at a multispecialty academic faculty practice organization. JAMA
  network open 2(3):e190554--e190554

\bibitem[{Devlin et~al.(2019)Devlin, Chang, Lee, and
  Toutanova}]{devlin2019bert}
Devlin J, Chang MW, Lee K, Toutanova K (2019) Bert: Pre-training of deep
  bidirectional transformers for language understanding. In: Proceedings of the
  2019 Conference of the North American Chapter of the Association for
  Computational Linguistics: Human Language Technologies, Volume 1 (Long and
  Short Papers), pp 4171--4186

\bibitem[{Finley et~al.(2018)Finley, Edwards, Robinson, Brenndoerfer, Sadoughi,
  Fone, Axtmann, Miller, and Suendermann-Oeft}]{finley2018automated}
Finley G, Edwards E, Robinson A, Brenndoerfer M, Sadoughi N, Fone J, Axtmann N,
  Miller M, Suendermann-Oeft D (2018) An automated medical scribe for
  documenting clinical encounters. In: Proceedings of the 2018 Conference of
  the North American Chapter of the Association for Computational Linguistics:
  Demonstrations, pp 11--15

\bibitem[{GAO(2018)}]{gao2018}
GAO (2018) Medical records: Fees and challenges associated with patients'
  access. United States Government Accountability Office, Report to
  Congressional Committees GAO-18-386

\bibitem[{Jeblee et~al.(2019)Jeblee, Khattak, Crampton, Mamdani, and
  Rudzicz}]{jeblee2019extracting}
Jeblee S, Khattak FK, Crampton N, Mamdani M, Rudzicz F (2019) Extracting
  relevant information from physician-patient dialogues for automated clinical
  note taking. In: Proceedings of the Tenth International Workshop on Health
  Text Mining and Information Analysis (LOUHI 2019), pp 65--74

\bibitem[{Kannan et~al.(2018)Kannan, Chen, Jaunzeikare, and
  Rajkomar}]{kannan2018semi}
Kannan A, Chen K, Jaunzeikare D, Rajkomar A (2018) Semi-supervised learning for
  information extraction from dialogue. In: Interspeech, pp 2077--2081

\bibitem[{Kessels(2003)}]{kessels2003patients}
Kessels RP (2003) Patients’ memory for medical information. Journal of the
  Royal Society of Medicine 96(5):219--222

\bibitem[{Kim et~al.(2019)Kim, Dalmia, and Metze}]{kim2019cross}
Kim S, Dalmia S, Metze F (2019) Cross-attention end-to-end asr for two-party
  conversations. arXiv preprint arXiv:190710726

\bibitem[{Kodish-Wachs et~al.(2018)Kodish-Wachs, Agassi, Kenny~III, and
  Overhage}]{kodish2018systematic}
Kodish-Wachs J, Agassi E, Kenny~III P, Overhage JM (2018) A systematic
  comparison of contemporary automatic speech recognition engines for
  conversational clinical speech. In: AMIA Annual Symposium Proceedings,
  American Medical Informatics Association, vol 2018, p 683

\bibitem[{Kumar(2016)}]{kumar2016burnout}
Kumar S (2016) Burnout and doctors: prevalence, prevention and intervention.
  In: Healthcare, Multidisciplinary Digital Publishing Institute, vol 4(3),
  p~37

\bibitem[{Leventhal(2018)}]{hcirajiv}
Leventhal R (2018) Physician burnout addressed: How one medical group is
  (virtually) progressing. Healthcare Innovation

\bibitem[{Liu et~al.(2005)Liu, Ma, Moore, Ganesan, and Nelson}]{liu2005rxnorm}
Liu S, Ma W, Moore R, Ganesan V, Nelson S (2005) Rxnorm: prescription for
  electronic drug information exchange. IT professional 7(5):17--23

\bibitem[{Liu et~al.(2019)Liu, Lim, Suhaimi, Tong, Ong, Ng, Lee, Macdonald,
  Ramasamy, Krishnaswamy et~al.}]{liu2019fast}
Liu Z, Lim H, Suhaimi NFA, Tong SC, Ong S, Ng A, Lee S, Macdonald MR, Ramasamy
  S, Krishnaswamy P, et~al. (2019) Fast prototyping a dialogue comprehension
  system for nurse-patient conversations on symptom monitoring. In: Proceedings
  of the 2019 Conference of the North American Chapter of the Association for
  Computational Linguistics: Human Language Technologies, Volume 2 (Industry
  Papers), pp 24--31

\bibitem[{McCann et~al.(2018)McCann, Keskar, Xiong, and
  Socher}]{mccann2018natural}
McCann B, Keskar NS, Xiong C, Socher R (2018) The natural language decathlon:
  Multitask learning as question answering. arXiv preprint arXiv:180608730

\bibitem[{Peters et~al.(2018)Peters, Neumann, Iyyer, Gardner, Clark, Lee, and
  Zettlemoyer}]{peters2018deep}
Peters ME, Neumann M, Iyyer M, Gardner M, Clark C, Lee K, Zettlemoyer L (2018)
  Deep contextualized word representations. In: Proceedings of NAACL-HLT, pp
  2227--2237

\bibitem[{See et~al.(2017)See, Liu, and Manning}]{see2017get}
See A, Liu PJ, Manning CD (2017) Get to the point: Summarization with
  pointer-generator networks. In: Proceedings of the 55th Annual Meeting of the
  Association for Computational Linguistics (Volume 1: Long Papers), pp
  1073--1083

\bibitem[{Vaswani et~al.(2017)Vaswani, Shazeer, Parmar, Uszkoreit, Jones,
  Gomez, Kaiser, and Polosukhin}]{vaswani2017attention}
Vaswani A, Shazeer N, Parmar N, Uszkoreit J, Jones L, Gomez AN, Kaiser {\L},
  Polosukhin I (2017) Attention is all you need. In: Advances in neural
  information processing systems, pp 5998--6008

\bibitem[{Xiong et~al.(2016)Xiong, Zhong, and Socher}]{xiong2016dynamic}
Xiong C, Zhong V, Socher R (2016) Dynamic coattention networks for question
  answering. arXiv preprint arXiv:161101604

\end{thebibliography}

\end{document}